\documentclass{article}
\usepackage{spconf,amsmath,epsfig,graphicx}
\usepackage[utf8]{inputenc}
\usepackage[T1]{fontenc}

\title{Concurrent segmentation and object detection CNNs for aircraft detection and identification in satellite images}

\name{Damien Grosgeorge, Maxime Arbelot, Alex Goupilleau, Tugdual Ceillier, Renaud Allioux}

\address{Earthcube, Paris, France}

\begin{document}
%
\maketitle
\begin{abstract}

Detecting and identifying objects in satellite images is a
very challenging task: objects of interest are often very small
and features can be difficult to recognize even using very high
resolution imagery. For most applications, this translates into
a trade-off between recall and precision. We present here a
dedicated method to detect and identify aircraft, combining
two very different convolutional neural networks (CNNs): a
segmentation model, based on a modified U-net architecture
\cite{ronneberger2015u}, and a detection model, based on the
RetinaNet architecture \cite{lin2017focal}. The results we present
show that this combination outperforms significantly each unitary
model, reducing drastically the false negative rate.

\end{abstract}
\begin{keywords}
CNNs, deep learning, segmentation, identification, aircraft, satellite images
\end{keywords}
\section{Introduction}
\label{sec:intro}

\begin{figure}[htb]
  \centering
  \begin{tabular}{cc}
    \includegraphics[width=.46\linewidth]{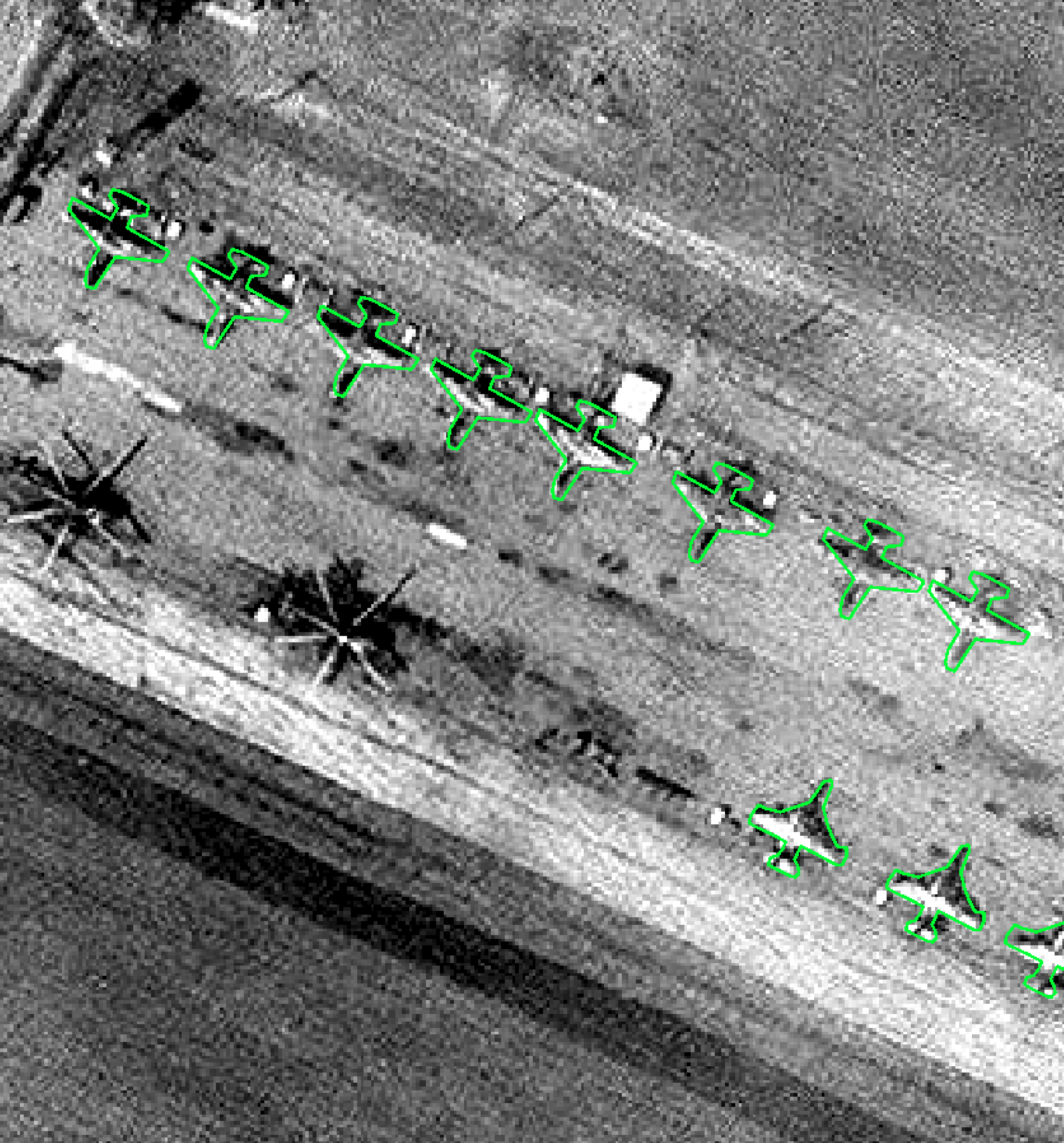}&
    \includegraphics[width=.412\linewidth]{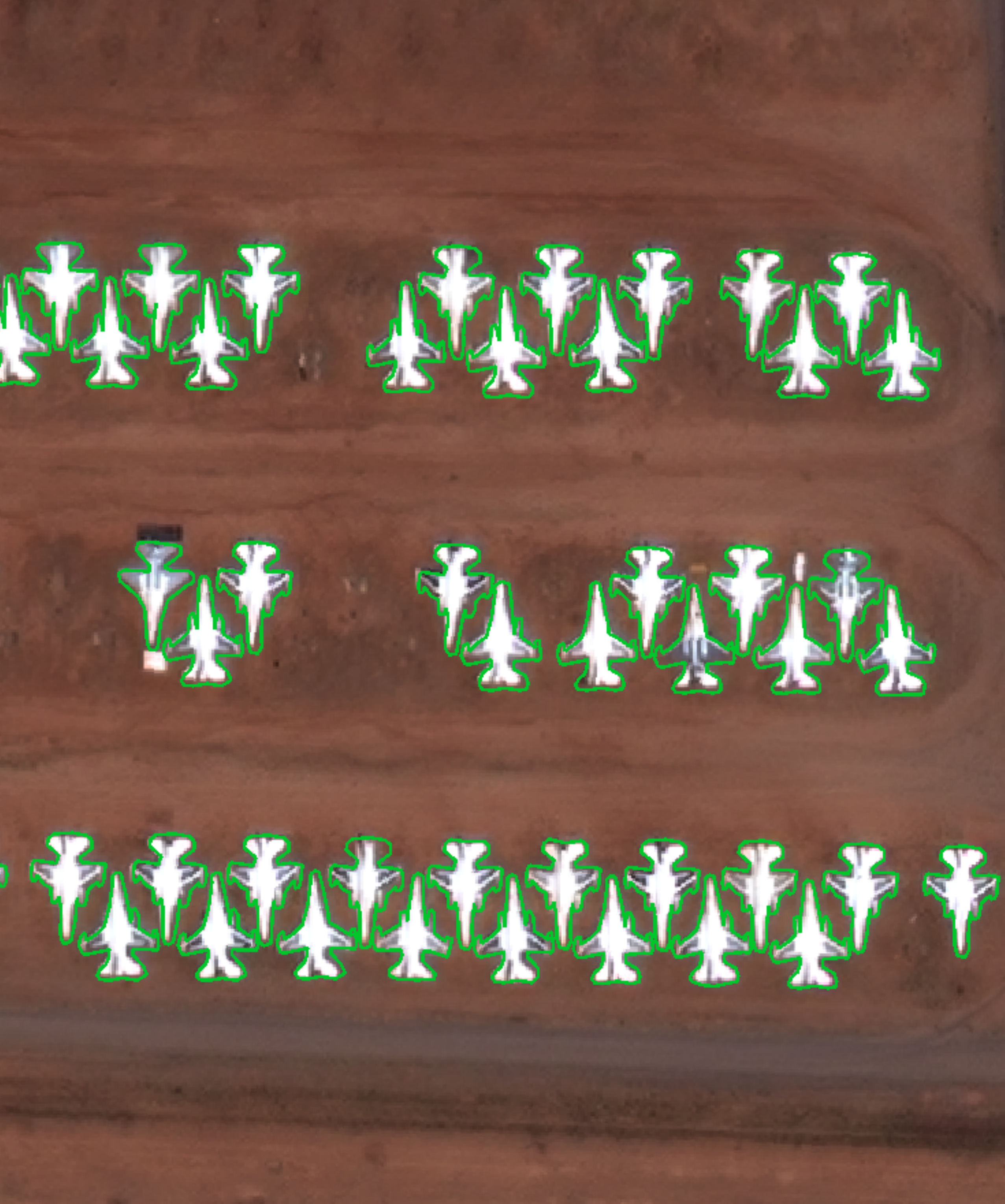}\\
    (a) Soukhoï Su-25 &
    (b) F-16 Fighting Falcon
  \end{tabular}
  \caption{Illustration of the data diversity (with ground truth).}
  \label{fig:hardexamples}
\end{figure}

The last decade has seen a huge increase of available high
resolution satellite images, which are used more and more for
surveillance tasks. When monitoring military sites, it is
necessary to automatically detect and identify objects of interest
to derive trends. In this domain, aircraft recognition is of
particular interest: each aircraft model has its own role, and a
variation in the number of a specific type of aircraft at a given
location can be a highly relevant insight. This recognition task
needs to be reliable to allow the automation of site analysis
-- in particular to derive alerts corresponding to unusual events.
Robustness to noise, shadows, illumination or ground texture
variation is challenging to obtain but mandatory for real-life
applications (see Fig.~\ref{fig:hardexamples}).

Nowadays, CNNs are considered as one of the best
techniques to analyse image content and are the most widely used
ML technique in computer vision applications. They have
recently produced the state-of-the-art results for image
recognition, segmentation and detection related
tasks \cite{liu2019recent}. A typical CNN architecture
is generally composed of alternate
layers of convolution and pooling (encoder) followed by a
decoder that can comprise one or more fully connected layers
(classification), a set of transpose convolutions
(segmentation) or some classification and regression branches
(object detection). The arrangement of the CNN components plays a
fundamental role in designing new architectures and thus in
achieving higher performances \cite{khan2019survey}.

For segmentation tasks, the U-net architecture has been widely
used since its creation by \cite{ronneberger2015u}. This
architecture allows a better reconstruction in the decoder by
using skip connections from the encoder (Fig. \ref{fig:unetarchi}).
Various improvements have been made in the literature considering
each CNN components \cite{khan2019survey}, but the global architecture
of the U-net is still one of the state-of-the-art architecture for
the segmentation task.

For detection tasks, two main categories have been
developed in the literature. The most well-known uses a
two-stages, proposal-driven mechanism: the first stage generates a
sparse set of candidate object locations and the second stage
classifies each candidate location either as one of the
foreground classes or as background using a CNN. One of the
most used two-stages model is the Faster-RCNN \cite{ren2015faster},
which has been considered as the state-of-the-art detector by
achieving top accuracy on the challenging COCO benchmark.
However, in the last few years, one-stage detectors, such as the
Feature Pyramid Network (FPN) \cite{lin2017feature}, have matched the
accuracy of the most complex two-stages detectors on the COCO
benchmark. In \cite{lin2017focal}, authors have identified that since
one-stage detectors are applied over a regular, dense sampling of
object locations, scales, and aspect ratios, then class
imbalance during training is the main obstacle impeding them from
achieving state-of-the-art accuracy. They thus proposed a new
loss function that eliminates this barrier (the focal loss) while
integrating improvements such as the FPN \cite{lin2017feature} in their model known as the RetinaNet \cite{lin2017focal}.

In this paper, we are looking for a dedicated and robust
approach to address the aircraft detection and identification
problems, that can be easily adapted to multiple applications.
We propose a hybrid solution based on different CNNs strategies:
a segmentation model based on the U-Net architecture
\cite{ronneberger2015u} for a better detection rate and an
object detection model based on the RetinaNet \cite{lin2017focal},
a fast one-stage detector, for identifying and improving the precision.
Section~\ref{sec:segretina} details this concurrent
approach while Section~\ref{sec:expres} presents results obtained on
high-resolution satellite images.

\section{Concurrent segmentation and object detection approach}
\label{sec:segretina}

In this section, we present the choices made in designing each
model considering the aircraft recognition problem, and how
they interact together. These choices are based on simple
observations: (i)~changing the paradigm of training modifies the
way features are learnt/extracted inside the model,
(ii)~segmentation models are really efficient but suffer from
bad separation and identification of objects, (iii)~in high-resolution
images from satellites, aircraft are of limited size. We also
based our choices on the latest developments in the field.

\subsection{Segmentation CNN}
\label{ssec:segcnn}
\begin{figure}[htb]
  \centering
  \includegraphics[width=.96\linewidth]{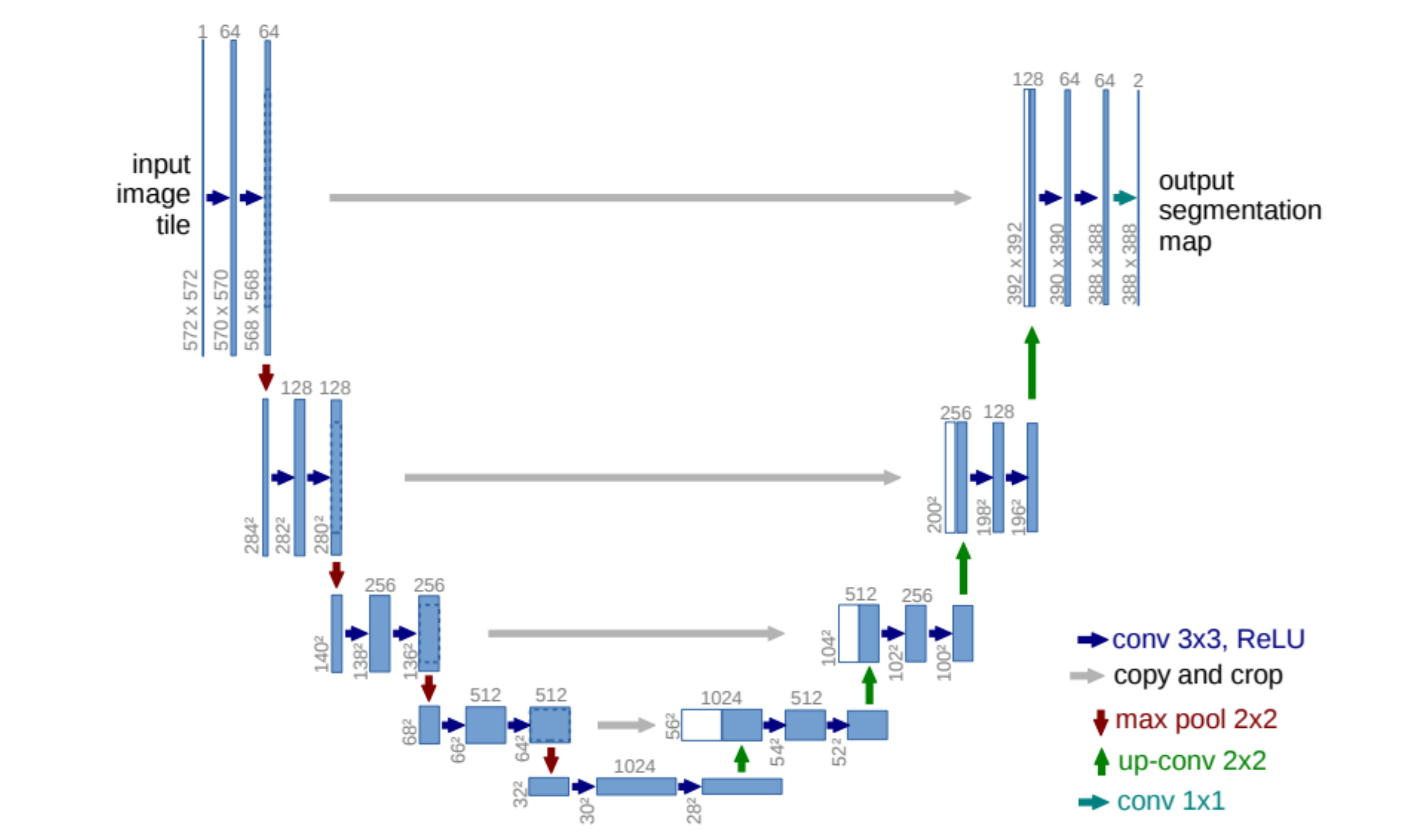}
\caption{Original U-Net architecture (from \cite{ronneberger2015u}).}
\label{fig:unetarchi}
\end{figure}

Our segmentation model is based on the U-net architecture, illustrated in Fig.~\ref{fig:unetarchi} (original architecture).

For our concurrent approach, the objective of this model is: (i)~to detect aircraft (without identification), (ii)~to have a very high recall (in particular for the location even if the delineation is of low quality), (iii)~to be robust to difficult cases (like occultation, shadow or noise). For that purpose, the U-net architecture has been updated:
\begin{itemize}
\setlength\itemsep{0em}
\item convolutionnal layers have been replaced by identity mapping (IM) blocks, as proposed by \cite{he2016identity}. It has been proven that this choice eases the training and the efficiency of deep networks;
\item maxpool layers have been replaced by convolutionnal layers with a stride of 2 (we reduce the spatial information while increasing the number of feature maps);
\item the depth and the width of the network have been set accordingly to the application: spatial information is only reduced twice (while doubling filters), the encoding is composed of $36$ IM blocks and the decoding of $8$ IM blocks (resp. $72$ and $16$ conv. layers).
\end{itemize}
Skip connections of the U-net are used for a better reconstruction of the prediction map.

\subsection{Object detection CNN}
\label{ssec:objcnn}
\begin{figure}[htb]
  \centering
  \includegraphics[width=.96\linewidth]{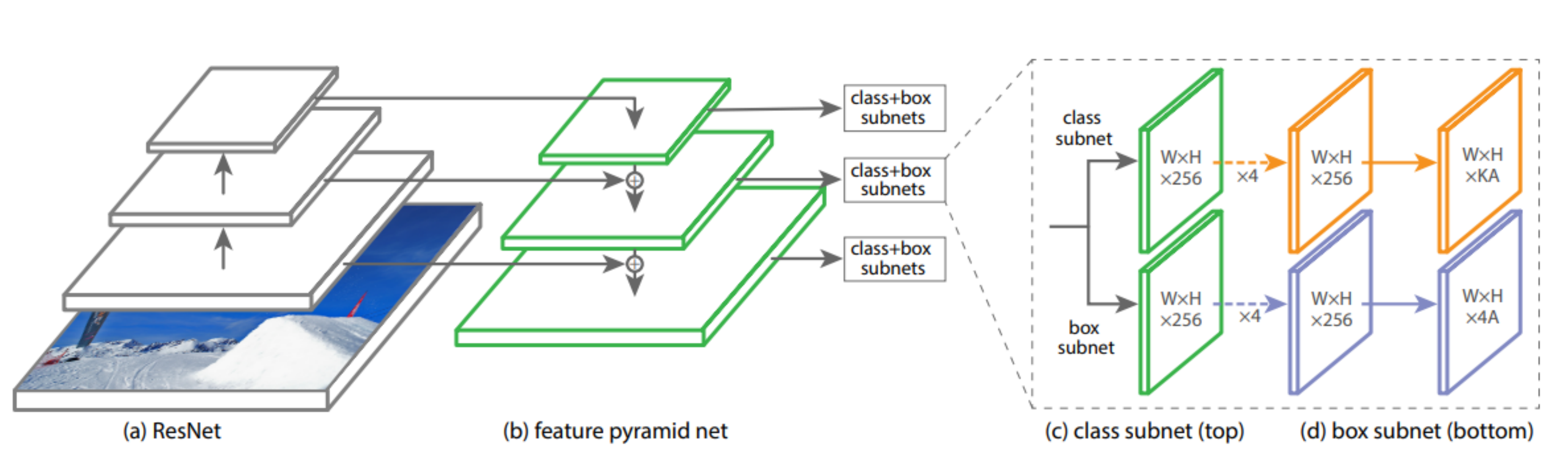}
\caption{Original RetinaNet architecture (from \cite{lin2017focal}).}
\label{fig:retinarchi}
\end{figure}

Our object detector is based on the RetinaNet architecture, illustrated by the Fig. \ref{fig:retinarchi}.

For our concurrent approach, the objective of this model is: (i)~to split the detected objects and (ii)~to correctly identify the objects. For that purpose, the RetinaNet architecture has been carefully set:
\begin{itemize}
\setlength\itemsep{0em}
\item one level has been added in the feature pyramid network \cite{lin2017feature} for a finest detection level (small objects);
\item the backbone of the RetinaNet model to extract features is a ResNet101;
\item a non maximal suppression (NMS) algorithm is used to remove duplicated results.
\end{itemize}
The focal loss proposed by \cite{lin2017focal} is used to address the foreground-background class imbalance encountered during the training.

\subsection{Concurrent approach}
\label{ssec:conccnn}

The training strategy of each model is different. Features of
the segmentation model are learnt on the aircraft objects, the
model is then good at localizing objects but is not designed for
separating or recognizing them. Features of the object
detection model are learnt on the finest aircraft identification (see
Section~\ref{ssec:data}), the model is then good at separating and
recognizing aircrafts but has a very low precision but a high recall.
The idea of our concurrent approach is to use these
complementary properties to improve detections. The process of the
system is sequential and can be summarized by the following
steps.
\begin{enumerate}
\setlength\itemsep{0em}
\item Apply the segmentation model on the unknown image to extract the prediction value for each pixel. This is the localization step.
\item Apply the object detector for each positive area of the localization step. This process can be iterative, considering how shift-invariant the object detection model is, by repeating: (i)~apply the detection model, (ii)~remove the detected objects from the segmentation map.
\item (optional) Study the remaining positive areas of the prediction map to increase the recall: add objects to the detected list considering size or distance to the detected aircraft.
\end{enumerate}

These steps allow the definition of several operating
modes considering the intrinsic qualities of the models:
parameters definition can yield a system dedicated to high
recall, to high precision or balanced.

\section{Experimental results}
\label{sec:expres}

\subsection{Data information}
\label{ssec:data}
\begin{figure}[htb]
\begin{minipage}[b]{1.0\linewidth}
  \centering
  \centerline{\epsfig{figure=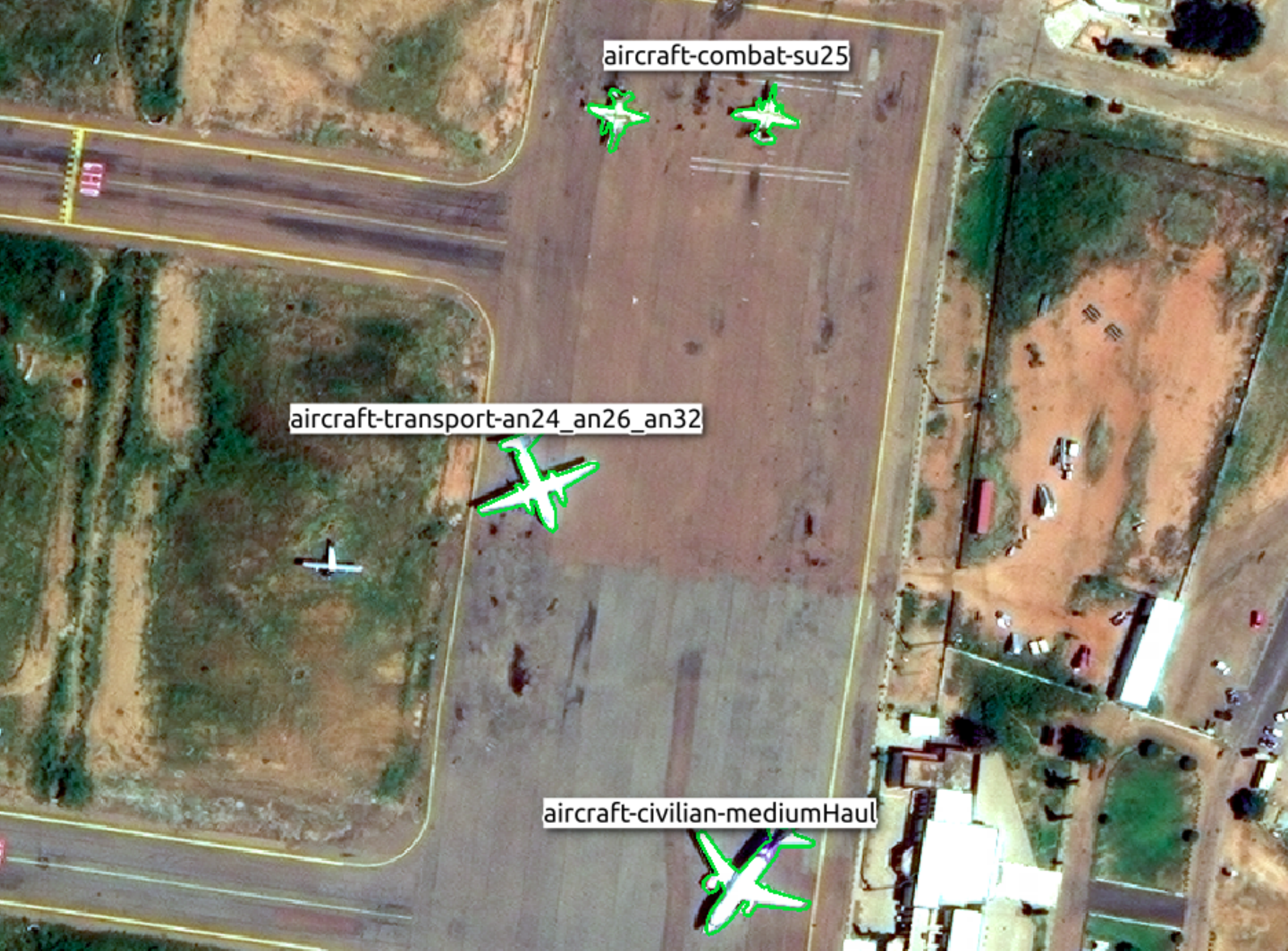,width=7.5cm}}
\end{minipage}
\caption{Example of the ground truth at level 3.}
\label{fig:gt_example}
\end{figure}

Our method has been applied to the aircraft recognition problem. Our datasets have three levels of aircraft identification: the first level is the type of the object (\emph{`aircraft`}), the second level represents the function of the aircraft (\emph{`bomber`}, \emph{`civilian`}, \emph{`combat`}, \emph{`drone`}, \emph{`special`} and \emph{`transport`}) and the third level is the aircraft identification. This last level is currently composed of 61 classes (for example \emph{`F-16`} Fighting Falcon is a third level of type \emph{`combat`} and Tupolev \emph{`Tu-95`} a third level of type \emph{`bomber`}). Fig.~\ref{fig:gt_example} shows an example of the ground truth at level 3.

Train and test datasets have been created using images from different satellites (resolution 30-50 cm). Train tiles are of size 512 pixels, with an overlap of 128 to improve shift invariance. The test dataset is composed of 30 satellite images at unknown locations (not seen during the training). Details of the datasets are given in Table~\ref{tab:dataset}.

\begin{table}
  \centering
  \begin{tabular}{c||c|c|c|c}

	Datasets & N img & N obj & N tiles & Area \\ [0.5ex] 
	\hline\hline
	Train - Seg & 9 984 & 122 479 & 105 206 & 51 166 \\

	Train - Obj & 10 179 & 128 422 & 361 843 & 49 905 \\
	\hline
	Test & 30 & 689 & - & 403 \\

  \end{tabular}
  \caption{Dataset information. Areas are in $km^{2}$.} \label{tab:dataset}
\end{table}

\subsection{Method parameterization}
\label{ssec:param}

The segmentation model has been trained using a weighted categorical cross-entropy loss:
\begin{equation}
wCE(y, \widehat{y}) = -\sum_{i=1}^{C}\alpha_i y_i \log\widehat{y_i}
\end{equation}
where $\widehat{y}$ is the prediction, $y$ the ground truth, $C$ the number of classes and $\alpha$ the median frequency balancing weights. These weights allow to balance the class distribution (compensate the high number of background pixels). ADAM optimizer has been used with an initial learning rate of $0.001$ (this one is decreased on plateau considering the validation loss).

The object detector has been trained using the focal loss \cite{lin2017focal} for the classification and the smooth L1 loss for the regression. We slightly increased the weighting of the classification compared to the regression (with a factor of $1.5$) and used the ADAM optimizer with an initial learning rate of $0.0004$. The NMS threshold has been set to $0.35$ (aircraft have a low overlap rate).

For both trainings, various data augmentations have been used to increase  model generalization: geometric transformations (flip, rotate) and radiometric transformations (grayscale, histogram equalization and normalization). For both models, different operating modes can be set by modifying two parameters: the prediction threshold and the minimum size. We empirically defined several modes, to balance recall and precision.

\subsection{Quantitative and qualitative results}
\label{ssec:quantquali}

\begin{figure}[htb]
  \centering
  \begin{tabular}{cc}
    \includegraphics[width=.92\linewidth]{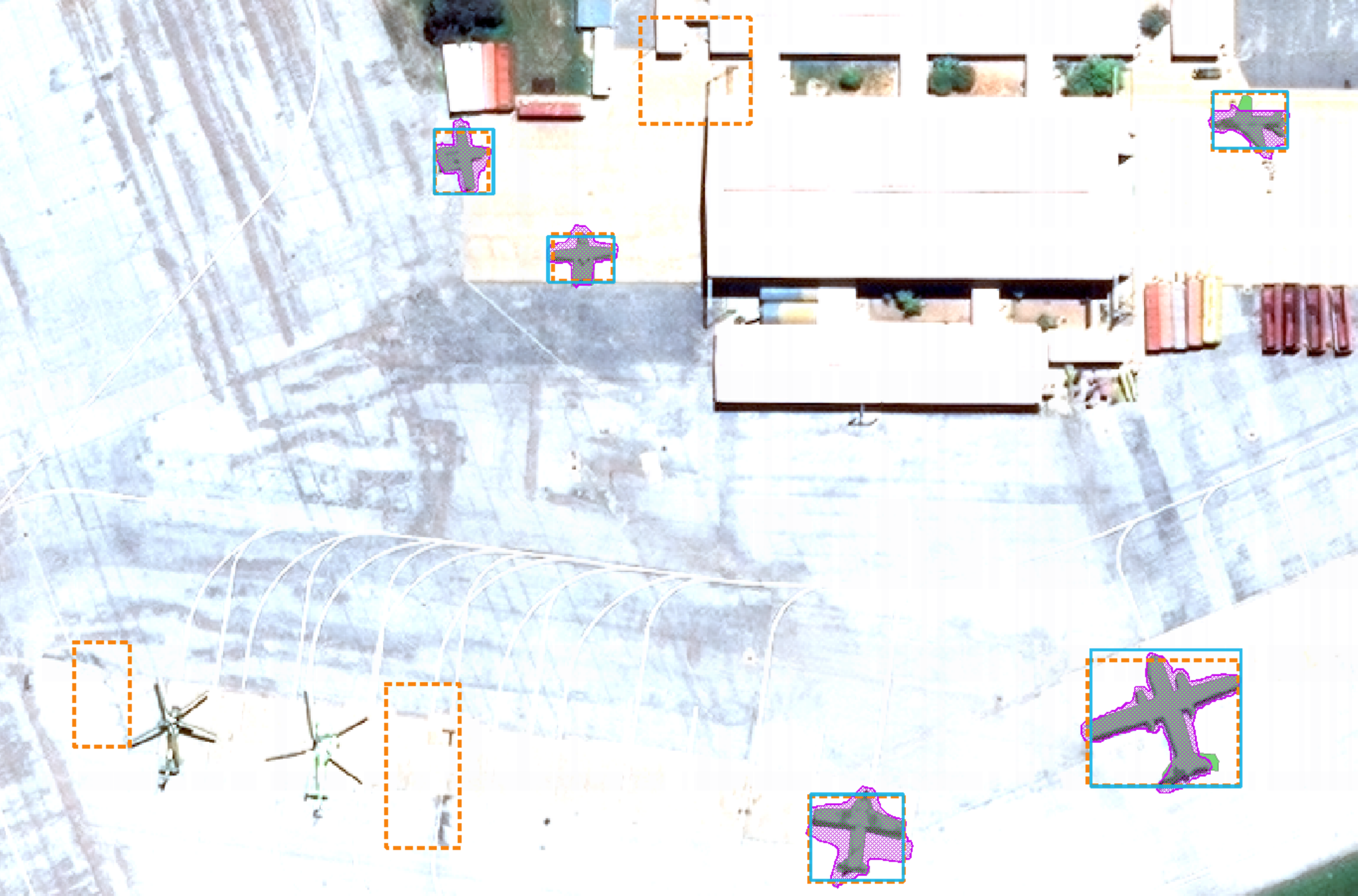}
  \end{tabular}
  \caption{Visual comparisons of ground truth (light green), the segmentation result (pink), the object detection model (orange dotted lines) and our method (light blue).}
\label{fig:comparisonsdetect}
\end{figure}

On our test dataset, we evaluated: (i)~the segmentation model alone (an overlap of 50\% is required to be considered as a detected aircraft), (ii)~the object detection alone and (iii)~our concurrent approach. Table \ref{tab:quantiresults} shows the detection results for each case, with two different modes: one balanced between recall and precision and one with a better recall. As expected, we can observe that our concurrent method allows to significantly increase the detection results compared to the segmentation model or the detection model alone: errors of each model are corrected by the other one to obtain better results (the false positives produced by the two models are not the same). This is illustrated in Fig.~\ref{fig:comparisonsdetect}: we can observe that false positives obtained with the object detection model are removed by our method.

\begin{table}[htb]
	\centering
	\begin{tabular}{ccccc} 
		{} &  \multicolumn{2}{c}{Balanced mode} & \multicolumn{2}{c}{Recall mode}\\
		{}   & R & P & R & P\\
		\hline\hline
		Segmentation & 0.91 & 0.78 & 0.95 & 0.5\\
		Object detection & 0.87 & 0.75 & 0.95 & 0.37\\
		\hline
		Our approach & \bf{0.95} & \bf{0.88} & \bf{0.96} & \bf{0.84}\\
	\end{tabular}
	\caption{Quantitative results of the aircraft detection on the test dataset (R: recall, P: precision).}
	\label{tab:quantiresults}
\end{table}

\begin{figure}[htb]
  \centering
  \begin{tabular}{cc}
	\multicolumn{2}{c}{\includegraphics[width=.9\linewidth]{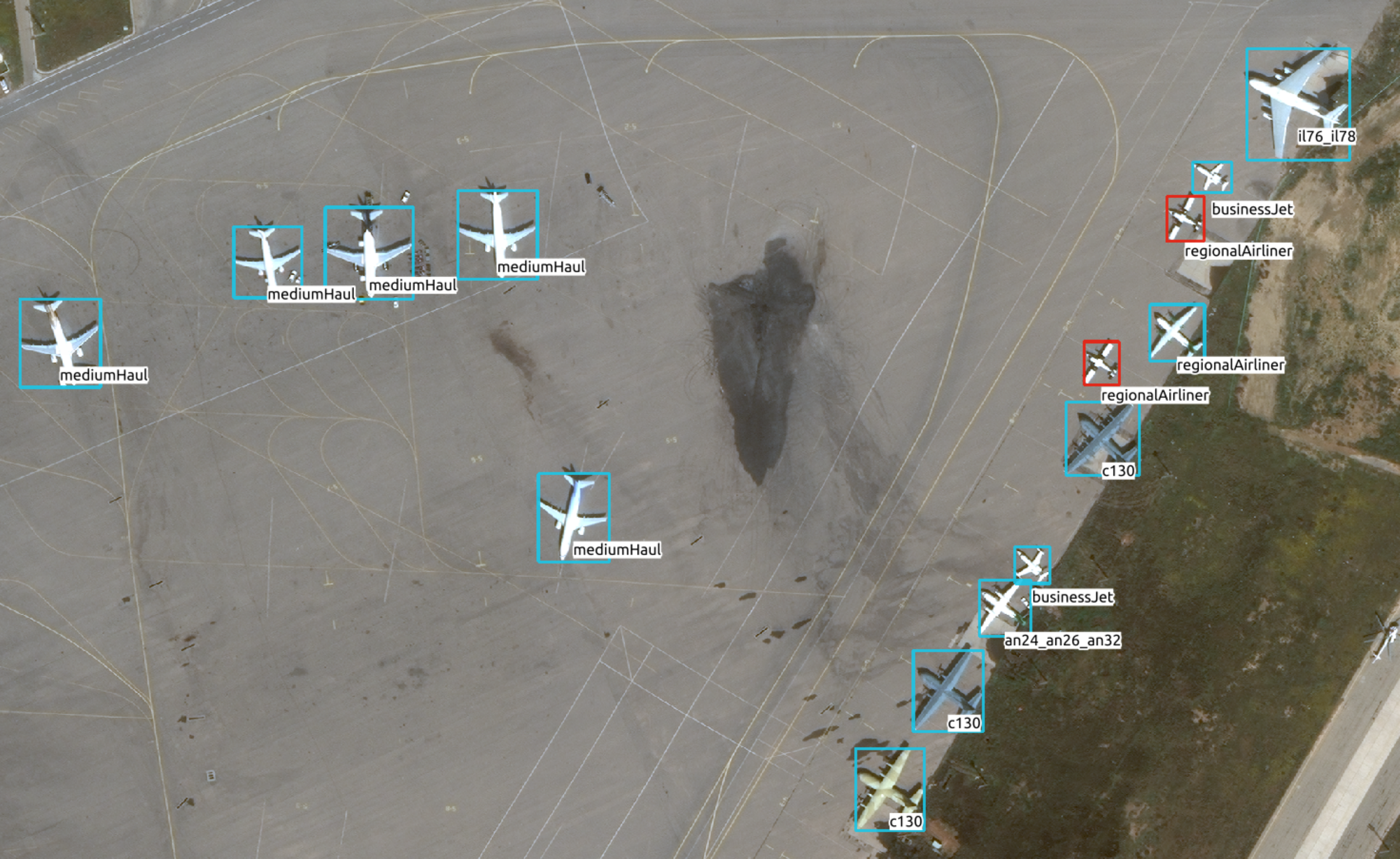}}\\
		\multicolumn{2}{c}{\includegraphics[width=.9\linewidth]{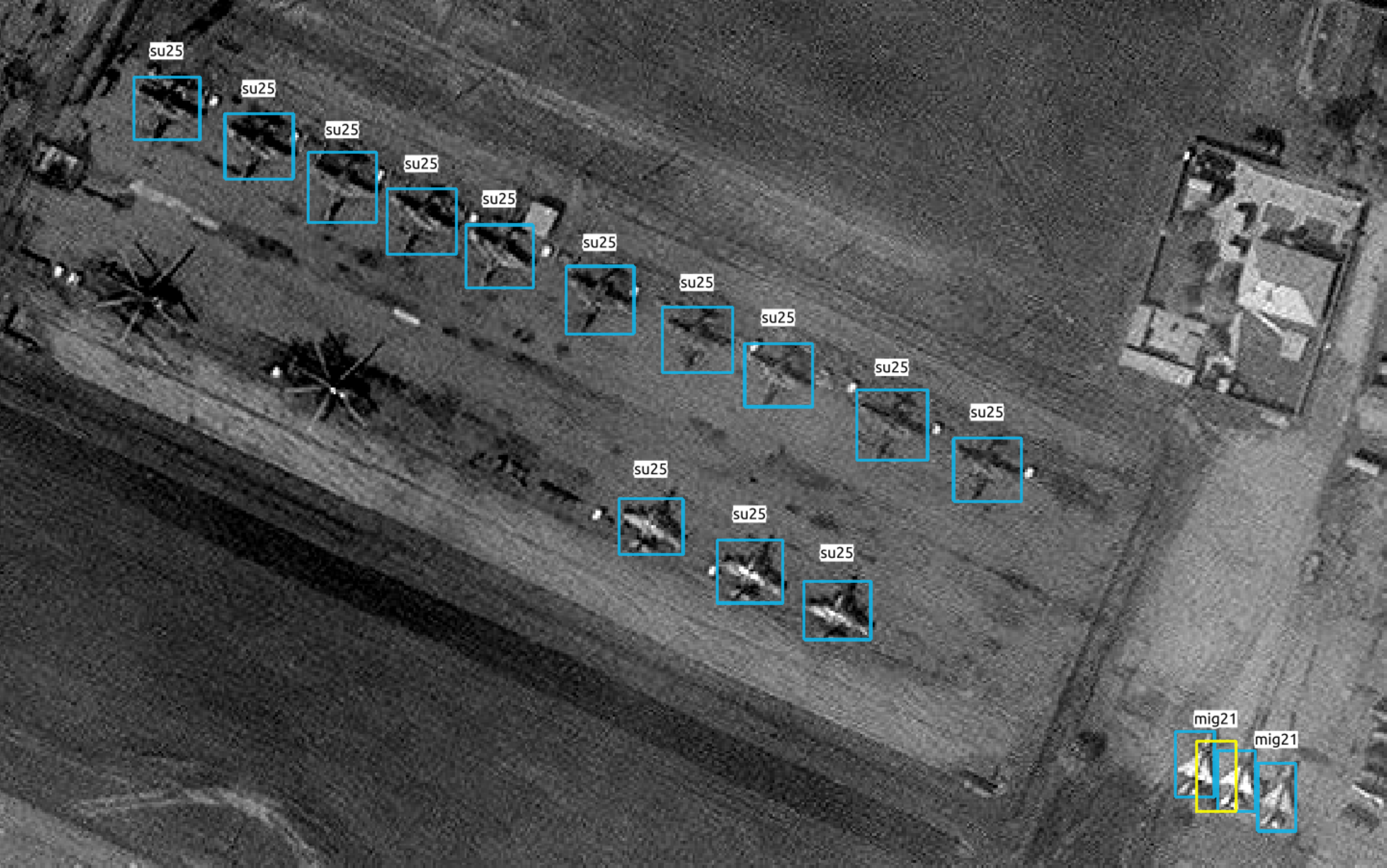}}\\  
  \end{tabular}    
\caption{Illustration of the aircraft classification. The good classifications are in blue, the wrong classifications in red, the false positives in yellow.}
\label{fig:classifvisu}
\end{figure}

On the same test dataset, we evaluated the identification of well-detected aircraft. The identification rate for the level~2 is $0.91$ and for the level~3 is $0.80$. Some errors happen because of the definition of some level~3 labels: regrouping different aircraft in the same class (for example \emph{small-aircraft}) can lead to confusion with combat aircraft. This can be seen in Fig.~\ref{fig:classifvisu}: the misclassified aircraft in the top image should have been assigned the \emph{small-aircraft} label.

\section{Conclusion and perspectives}
\label{sec:concl}

In this work, we developed a concurrent method combining
two CNNs: a segmentation model and a detection model. We
have shown that this combination allows to significantly
improve aircraft detection results (very low false detection rate
and high rate of good identification). In the future, we plan
on: (i)~refining our level~3 dataset in order to avoid some
identification confusions and (ii)~designing an all-in-one model
integrating level~1 and level~3 features in the same architecture.


\bibliographystyle{IEEEbib}
\bibliography{biblio}

\end{document}